\documentclass{article}

\usepackage{PRIMEarxiv}

\usepackage[utf8]{inputenc} 
\usepackage[T1]{fontenc}    
\usepackage{hyperref}       
\usepackage{url}            
\usepackage{booktabs}       
\usepackage{amsfonts}       
\usepackage{nicefrac}       
\usepackage{microtype}      
\usepackage{lipsum}
\usepackage{fancyhdr}       
\usepackage{graphicx}       
\graphicspath{{media/}}     

\title{Artificial Intelligence for Personalized Prediction of Alzheimer's Disease Progression: A Survey of Methods, Data Challenges, and Future Directions
\thanks{\textit{\underline{Citation}}: 
\textbf{Hancerliogullari Koksalmis, Soykan, Brattain and Huang. Artificial Intelligence for Personalized Prediction of Alzheimer's Disease Progression: A Survey of Methods, Data Challenges, and Future Directions.  } 
}}

\author{
  Gulsah Hancerliogullari Koksalmis \\
  Department of Industrial Engineering and Management Systems \\
  University of Central Florida \\
  Orlando, FL, USA\\
  \texttt{gulsah.hancerliogullarikoksalmis@ucf.edu} \\
   \And
  Bulent Soykan \\
  Institute for Simulation and Training \\
  University of Central Florida \\
  Orlando, FL, USA\\
  \texttt{bulent.soykan@ucf.edu} \\
     \And
  Laura J. Brattain \\
  Department of Internal Medicine \\
  University of Central Florida \\
  Orlando, FL, USA\\
  \texttt{laura.brattain@ucf.edu} \\
     \And
  Hsin-Hsiung Huang \\
  Department of Statistics and Data Science \\
  University of Central Florida \\
  Orlando, FL, USA\\
  \texttt{hsin-hsiung.huang@ucf.edu} \\
}

\begin{document}
\maketitle

\begin{abstract}
Alzheimer's Disease (AD) is marked by significant inter-individual variability in its progression, complicating accurate prognosis and personalized care planning. This heterogeneity underscores the critical need for predictive models capable of forecasting patient-specific disease trajectories. Artificial Intelligence (AI) offers powerful tools to address this challenge by analyzing complex, multi-modal, and longitudinal patient data. This paper provides a comprehensive survey of AI methodologies applied to personalized AD progression prediction. We review key approaches including state-space models for capturing temporal dynamics, deep learning techniques like Recurrent Neural Networks for sequence modeling, Graph Neural Networks (GNNs) for leveraging network structures, and the emerging concept of AI-driven digital twins for individualized simulation. Recognizing that data limitations often impede progress, we examine common challenges such as high dimensionality, missing data, and dataset imbalance. We further discuss AI-driven mitigation strategies, with a specific focus on synthetic data generation using Variational Autoencoders (VAEs) and Generative Adversarial Networks (GANs) to augment and balance datasets. The survey synthesizes the strengths and limitations of current approaches, emphasizing the trend towards multimodal integration and the persistent need for model interpretability and generalizability. Finally, we identify critical open challenges, including robust external validation, clinical integration, and ethical considerations, and outline promising future research directions such as hybrid models, causal inference, and federated learning. This review aims to consolidate current knowledge and guide future efforts in developing clinically relevant AI tools for personalized AD prognostication.

\end{abstract}


\section{Introduction}

Alzheimer's Disease (AD) stands as a formidable neurodegenerative disorder and represents the most common cause of dementia worldwide \cite{breijyeh2020comprehensive}. It constitutes a significant and escalating public health challenge, driven largely by aging global populations. The disease is defined by its insidious onset and progressive nature, involving the gradual loss of neurons and synaptic connections in the brain. This deterioration typically manifests initially as mild memory loss but inexorably advances, leading to severe impairments in cognitive functions, including thinking, reasoning, and judgment, as well as changes in behavior and personality, ultimately resulting in a complete loss of functional independence. Beyond the profound personal toll on affected individuals and their families, AD imposes an immense economic and social burden. This includes substantial direct healthcare costs associated with diagnosis and treatment, extensive costs related to long-term formal and informal care, and significant indirect costs stemming from lost productivity of both patients and caregivers. The social fabric is also heavily impacted, contributing to caregiver stress, reduced quality of life for families, and increased strain on community support systems.

A significant challenge in managing AD stems from its remarkable heterogeneity; the rate and pattern of cognitive decline, symptom manifestation, and pathological changes vary considerably from one individual to another. This high inter-patient variability complicates clinical practice and research efforts. Current diagnostic frameworks, prognostic tools, and treatment strategies often rely on generalized, population-level models or staging systems. These "one-size-fits-all" approaches inherently struggle to capture the unique trajectory of each patient. Consequently, they can lead to inaccuracies in predicting future cognitive status, suboptimal timing of interventions, difficulties in stratifying patients for clinical trials, and challenges in providing tailored care and counsel to patients and families. There is, therefore, a critical and unmet need for accurate and robust predictive models capable of forecasting disease progression on an individual basis, taking into account each patient's specific clinical, biological, and demographic characteristics.

Addressing the challenge of predicting individualized AD progression requires tools capable of handling the inherent complexity and high dimensionality of relevant patient data. Artificial Intelligence (AI) and its subfield, Machine Learning (ML), offer a powerful potential solution space. These computational methods excel at identifying intricate patterns and relationships within large, multifaceted datasets that often elude traditional statistical approaches. By integrating diverse data sources – including neuroimaging, biomarker measurements, genetic information, clinical assessments, and demographic details – AI and ML algorithms can learn sophisticated representations of disease dynamics. This capability holds significant promise for developing models that can capture the subtle nuances of individual patient trajectories, moving beyond population averages to generate personalized forecasts of how AD is likely to progress for a specific person.

The primary objective of this paper is to provide a comprehensive survey and synthesis of existing AI methodologies specifically developed and applied for the purpose of personalized AD progression prediction. Our focus is strictly on approaches that aim to forecast individual patient trajectories, rather than general AD classification or diagnosis. We will systematically review and categorize key AI techniques leveraged for this task, including state-space and sequence models designed to capture temporal dynamics, Graph Neural Networks (GNNs) utilized for modeling relationships in patient or brain data, and the emerging concept of AI-driven digital twins for personalized simulation. Furthermore, given the practical challenges associated with AD datasets, we will also examine relevant AI-based strategies for mitigating data issues, with a particular focus on synthetic data generation using Variational Autoencoders (VAEs) and Generative Adversarial Networks (GANs) to address class imbalance in progression rates. This review aims to consolidate current knowledge, highlight methodological trends, and identify gaps in the field of personalized AD forecasting using AI.

The remainder of this paper is organized as follows: Section 2 provides essential background information on AD progression dynamics, common data modalities used for prediction, and the inherent challenges associated with disease heterogeneity and personalization. Section 3 delves into the core AI methodologies employed for personalized AD progression prediction, detailing approaches such as state-space models, GNNs, and AI-driven digital twins. Section 4 specifically addresses common data challenges encountered in this domain, with a focus on techniques like synthetic data generation using VAEs and GANs to handle class imbalance. Finally, Section 5 presents a discussion synthesizing the reviewed literature, highlights current limitations and open research questions, proposes future directions, and offers concluding remarks on the state and potential of AI in personalized AD forecasting.

\section{Bibliometric Overview: Trends in AI-Driven Personalized Prediction for Cognitive Decline}

In order to contextualize the growing interest in applying artificial intelligence and to explore the current landscape of research focused on AI applications in cognitive health, this subsection adopts a scientometric analysis which uncovers trends, publication growth and advancements in the field to reveal core research clusters and thematic connections within the body of literature.

\begin{figure}[h]
    \centering
    \includegraphics[scale=0.5]{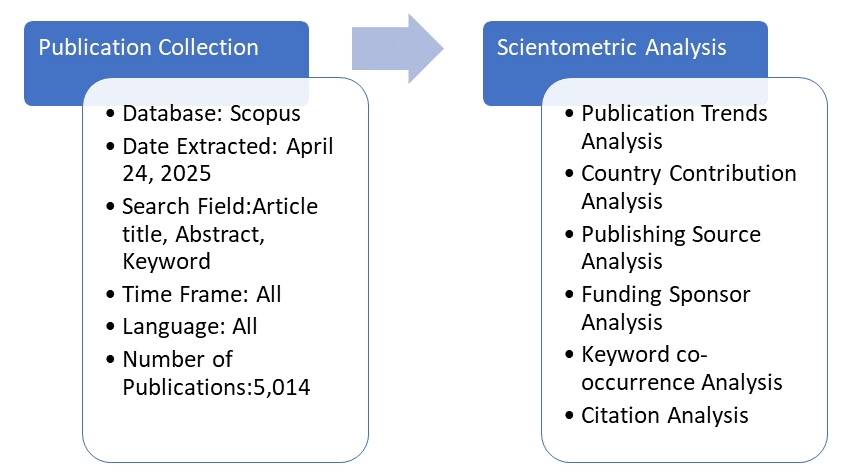} 
    \caption{Flow chart illustrating the scientometric analysis process. The process involved publication collection from the Scopus database using specified criteria (date, search fields, time frame, language) resulting in 5,014 publications, followed by scientometric analysis including descriptive, co-occurrence, and citation analyses. This figure outlines the systematic methodology employed for the bibliometric review, ensuring transparency regarding data sourcing and analytical techniques.}
    \label{fig:Figure 1}
\end{figure}

The publications (i.e., bibliographic records) are collected from the Scopus database, recognized for its comprehensive coverage. As of March 24, 2025, there have been 5,014 publications. A 37-year timeframe is selected, beginning in 1988-the year the earliest publication related to AI-driven cognitive heath research appeared-in order to capture the full scope of the literature on the topic up to March 24, 2025. The search term used in Scopus is (TITLE-ABS-KEY ("cognitive decline"OR"dementia"OR"alzheimer's"OR"aging") AND TITLE-ABS-KEY ("artificial intelligence"OR"AI"OR"chatgbt"OR"digital twin"OR"deep learning"OR"machine learning"OR"large language model"OR"LLM") AND TITLE-ABS-KEY ("MRI"OR"FMRI"OR"PET"OR"EEG") ). The publication collection process is illustrated in Figure~\ref{fig:Figure 1}, which presents a flow chart.  

\begin{figure}[htbp]
    \centering
    \includegraphics[scale=0.5]{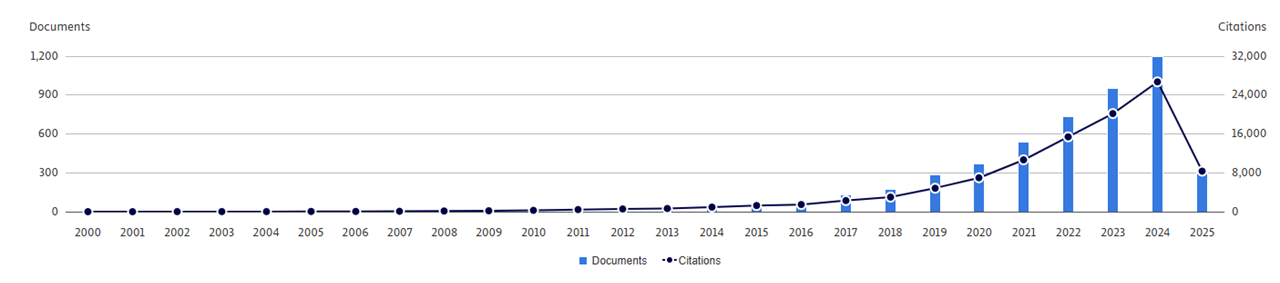} 
    \caption{Yearly publication volume (bars) and citation trends (line graph) from 2000 to 2025 (data as of early 2025). The plot clearly shows a marked acceleration in both the number of publications and citations starting around 2015. Research activity and scholarly impact concerning AI applications in cognitive decline and neuroimaging have experienced exponential growth in the last decade, indicating rapidly increasing interest and relevance.}
    \label{fig:Figure 2}
\end{figure}

In this study, quantitative scientometric techniques are applied utilizing VOSviewer and Microsoft Excel. VOSviewer enabled the exploration and mapping of scientometric data by employing a clustering technique grounded in modularity optimization \cite{van2010software,damar2024bibliometric,koksalmis2023artificial}. Meanwhile, Microsoft Excel supported the creation of descriptive charts and tables. This research examined the progression of yearly publication and citation patterns spanning from 1988 to 2025. Figure ~\ref{fig:Figure 2} depicts the yearly publication and citation trends.

\begin{figure}[htbp]
    \centering
    \includegraphics[scale=0.5]{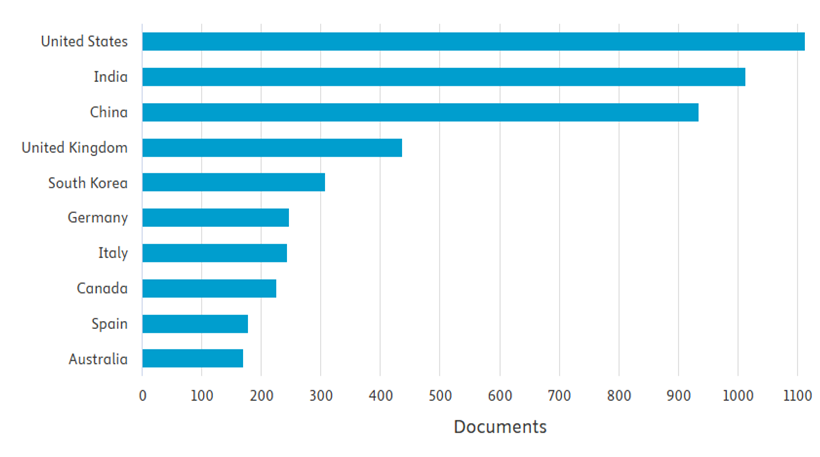} 
    \caption{Top ten countries contributing to publications based on the search criteria. The United States leads significantly in publication volume, followed by India and China. This figure identifies the major global contributors to research in this field, highlighting the leading nations driving advancements in AI for cognitive health and neuroimaging.}
    \label{fig:Figure 3}
\end{figure}

The results showed that between 1988 and 2014, the trend remained relatively stable. However, from 2015 onward, there was a notable and continuous increase. Notably, the growth in publication volume closely corresponded with the upward trajectory of citation counts. The data demonstrates a consistent rise in overall citations over time. This increasing trend highlights the subject's sustained relevance in scholarly research. Figure~\ref{fig:Figure 3} examines the top ten countries contributing to the. The United States leads with 1136 publications, followed by India, China, United Kingdom and South Korea. 

\begin{figure}[htbp]
    \centering
    \includegraphics[scale=0.5]{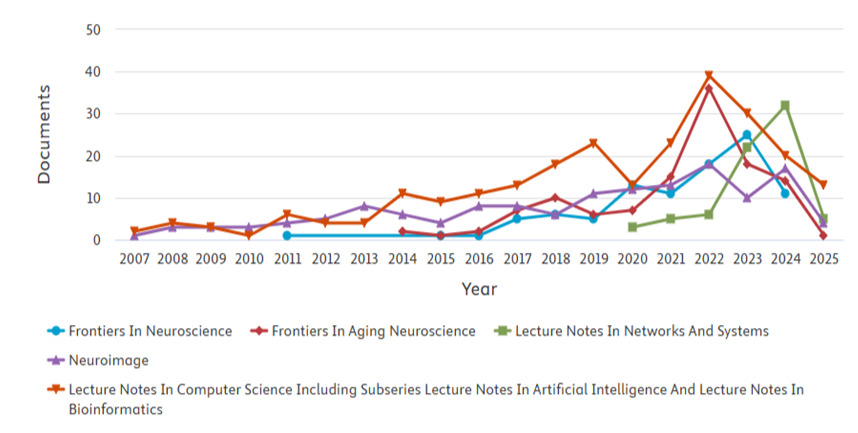} 
    \caption{Publication trends over time (2007-2025) for the top five publishing sources identified in the search. 'Lecture Notes In Computer Science...' and 'Frontiers' journals are prominent outlets, showing varying publication trajectories over the period. This visualization pinpoints the core academic venues where research at the intersection of AI, cognitive decline, and neuroimaging is most frequently published, guiding readers to key sources.}
    \label{fig:Figure 4}
\end{figure}

Figure~\ref{fig:Figure 4} displays the top five sources publishing research on the topic. Lecture Notes In Computer Science Including Subseries Lecture Notes In Artificial Intelligence And Lecture Notes In Bioinformatics leads the list with 247 publications, followed by Neuroimage, Frontiers in Aging Neuroscience, Frontiers in Neuroscience and Lecture Notes in Networks and Systems. These journals represent the primary outlets for advancing research in this field. Therefore, scholars seeking the latest and most impactful studies in this domain may consider beginning their search with these leading publication venues.

\begin{figure}[htbp]
    \centering
    \includegraphics[scale=0.5]{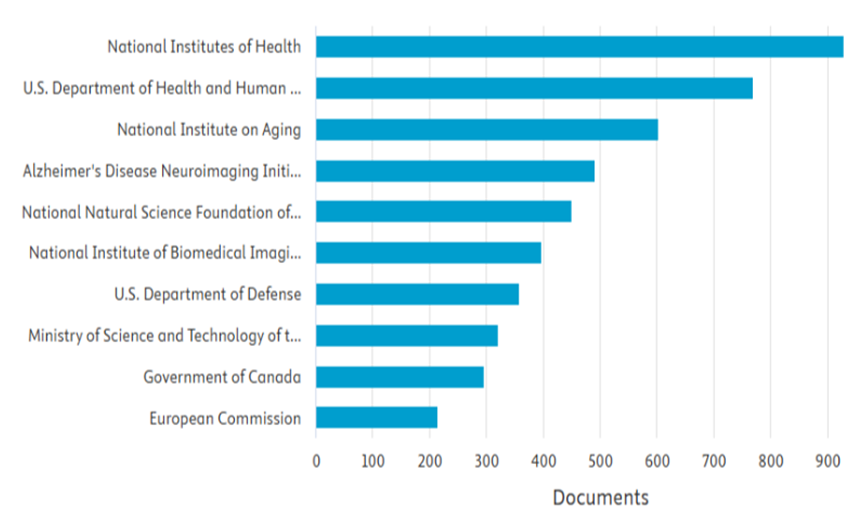} 
    \caption{Top ten funding sponsors supporting research in the analyzed field, ranked by the number of publications acknowledging their support. The National Institutes of Health (NIH) and other U.S. governmental bodies (HHS, NIA) are the most prominent funders.  This highlights the critical role of major funding agencies, particularly the NIH, in enabling and shaping the research landscape for AI applications in cognitive health.}
    \label{fig:Figure 5}
\end{figure}

Figure~\ref{fig:Figure 5} presents the top ten funding sponsors for research on the topic. National Institutes of Health is the leading sponsor, supporting 943 publications, followed by U.S. Department of Health and Human Services, National Institute on Aging, Alzheimer's Disease Neuroimaging Initiative (ADNI). These organizations play a crucial role in funding and advancing research in this area. Accordingly, researchers aiming to align their work with well-supported and high-impact initiatives may benefit from exploring funding opportunities through these key organizations.

\begin{figure}[htbp]
    \centering
    \includegraphics[scale=0.5]{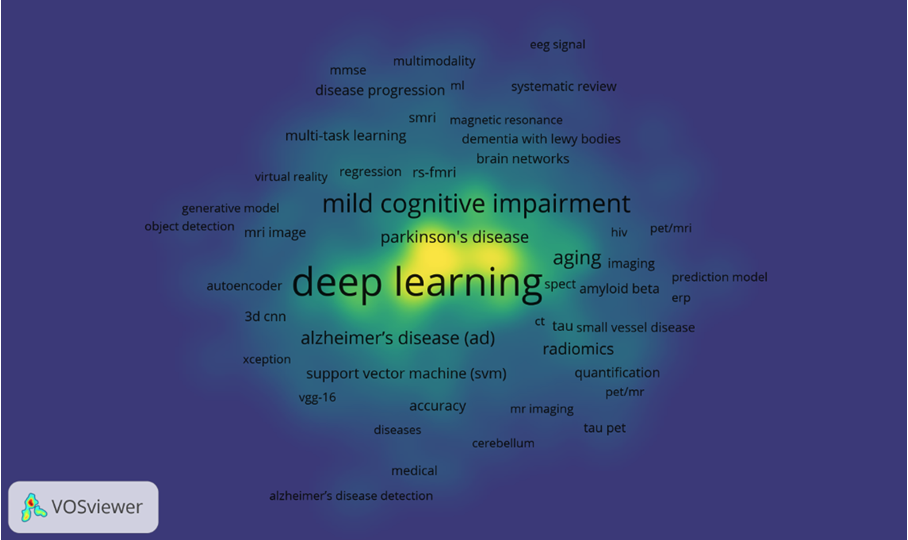} 
    \caption{VOSviewer density map illustrating the co-occurrence network of author keywords appearing at least five times. The central and brightest areas highlight 'deep learning', 'alzheimer's disease (ad)', and 'mild cognitive impairment' as core, highly interconnected themes. This map reveals the thematic structure of the research field, identifying central topics and their connections to specific AI techniques, pathologies, imaging modalities, and related conditions.}
    \label{fig:Figure 6}
\end{figure}

A co-occurrence analysis is performed on author keywords that appeared more than five times, as visualized via VOSviewer in Figure~\ref{fig:Figure 6}. Out of a total of 7,231keywords, 522 met the inclusion criteria for analysis, resulting in the formation of 17 distinct clusters.

\begin{figure}[htbp]
    \centering
    \includegraphics[scale=0.5]{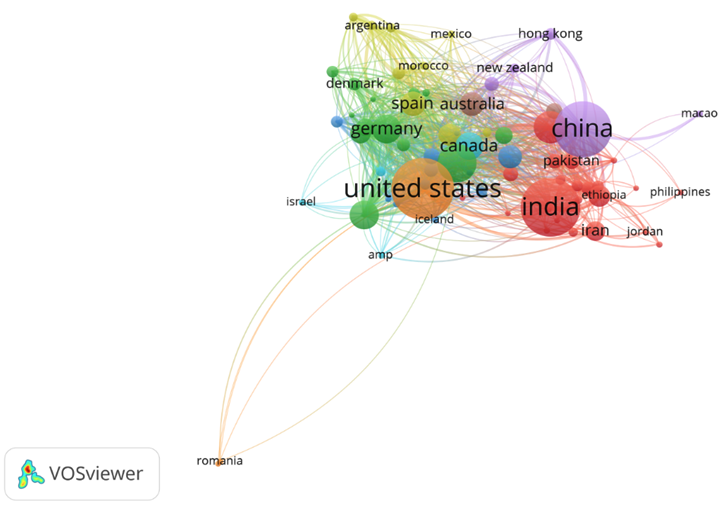} 
    \caption{VOSviewer network map visualizing international co-authorship collaborations among countries with at least five publications. Node size and link thickness indicate collaboration strength and frequency. The United States, China, India, and several European nations emerge as major hubs in the global collaborative network. This figure illustrates the structure of international research collaboration in the field, identifying key countries acting as central nodes in the co-authorship network.}
    \label{fig:Figure 7}
\end{figure}

A co-authorship analysis was conducted to examine collaborations in digital twin research for cognitive health, focusing on countries with at least five publications. Among 140 nations, 73 met this criterion, forming eight distinct clusters, as depicted via VOSviewer in Figure~\ref{fig:Figure 7}. The findings reveal that all countries are interconnected within the collaborative network. The most influential nations, based on Total Link Strength (TLS) and citation volume, include the United States, United Kingdom and China.

These findings reflect significant institutions, governments, encouraging chances for collaboration and efficient resource allocation.  It provides an analysis of the evolving trends in this discipline over the years, shedding insight into the growth and changes in both publications and citations. This analysis adds to our understanding of how related research is emerging. These trends suggest an evolving focus toward AI-based modeling approaches in AD research. The results affirm the growing momentum of AI-based methodologies aimed at improving personalized prognosis and treatment planning for Alzheimer’s Disease, thereby supporting the need for a structured survey of current modeling strategies.

\section{AI Methodologies for Personalized AD Progression Prediction}

\begin{figure}[htbp]
    \centering
    \includegraphics[scale=0.5]{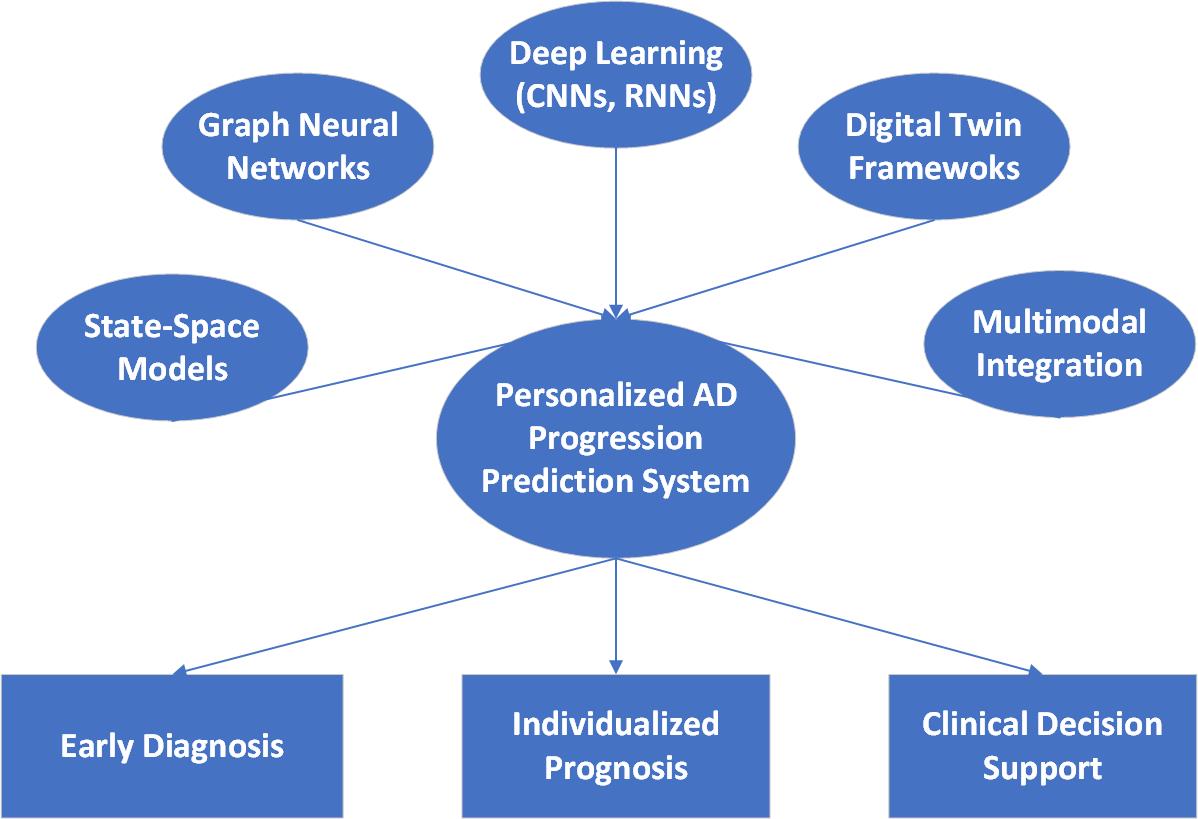} 
    \caption{Conceptual framework illustrating how diverse AI methodologies (State-Space Models, Graph Neural Networks, Deep Learning, Digital Twin Frameworks, Multimodal Integration) contribute to a centralized Personalized AD Progression Prediction System. This system, in turn, supports key clinical applications: Early Diagnosis, Individualized Prognosis, and Clinical Decision Support. This diagram highlights the synergistic potential of combining various advanced AI techniques within a unified system to produce actionable insights across different stages of AD clinical management.}
    \label{fig:Figure 8}
\end{figure}

This section examines several AI approaches that have been proposed to predict the progression of AD, with a focus on personalized prediction models. Predicting disease progression precisely is important to provide appropriate interventions, improve patient care, and evaluate the efficiency of clinical treatments. State-of-the-art AI techniques, which contain state-space models,GNNs, and AI-driven digital twins, that have presented encouraging output are reviewed. To provide a high-level overview of how these advanced AI methodologies interact to support early diagnosis, individualized prognosis, and clinical decision-making in Alzheimer's disease, a conceptual framework is presented in Figure~\ref{fig:Figure 8}.

AD is a condition that gradually creeps up on individuals, silently affecting their brain, causing memory loss, making it increasingly difficult to think clearly, and having a harder time doing even the simplest tasks. It's not just tough for the person with the disease, it's also very hard on doctors and caregivers who are trying to help \cite{jack2018nia}. A big part of managing AD is to determine how it is going to progress. If doctors can get a sense of where it is headed, they can step in sooner and personalize treatment to what that person needs, which usually leads to better results \cite{dubois2016preclinical}. The problem is that AD doesn't appear the same for everyone. Each person's disease progresses at a different rate, symptoms might appear at different times, and treatment outcomes can differ greatly. It is challenging to rely on generic models that apply to everyone because different people may experience different symptoms at different times and rates \cite{blennow2015clinical}.

Historically, clinicians have relied on cognitive assessments to detect and monitor decline. Key examples include the Mini-Mental State Examination (MMSE), introduced by Folstein et al. \cite{folstein1975mini}, and the Alzheimer’s Disease Assessment Scale \textit{Cognitive subscale} (ADAS-Cog), developed in the 1980s \cite{Rosen1984}. These established essential clinical benchmarks for evaluating cognitive impairment and dementia severity, often supplemented by measures like the Clinical Dementia Rating sum of boxes (CDR-SB) and functional assessments (e.g., FAQ). While these instruments remain in use, their utility is limited by their inability to detect impairment before significant neurodegeneration occurs. Conventional diagnostic approaches---including cognitive testing, clinical evaluations, and neuroimaging---typically identify AD at a relatively advanced stage, when the underlying pathology may already be irreversible. This reality underscores the urgent need for methodologies capable of detecting AD at earlier, preclinical stages. Early diagnosis offers critical benefits: it enables patients and caregivers to plan proactively, initiate preventive strategies, and potentially slow the disease’s progression \cite{nagarajan2025comprehensive}. Given global demographic trends and rising prevalence rates, the demand for precise and timely identification of AD is becoming increasingly urgent. This pressing need has driven the exploration of advanced AI-based modeling approaches for understanding and predicting cognitive decline.

To address this, some recent breakthroughs in AI are making a big difference. From brain scans to genetic data, patient health records, and clinical evaluations, artificial intelligence is helping construct more customized models to forecast how AD will play out for each individual \cite{liu2014multimodal}. Deep learning (DL), GNNs, even digital twins are showing genuine promise in helping doctors understand AD better and make wiser decisions \cite{lecun2015deep}. AI distinguishes AD research mostly in its capacity to detect minute, non-linear changes in the course of the illness. Given Alzheimer's involves complex interactions between many biological systems, conventional methods sometimes find it difficult to spot these minor trends \cite{chetelat2005mapping}. AI can reveal these underlying trends by sorting through massive and diverse datasets, allowing for more tailored projections that better understand the disease's expected trajectory.

Unlike traditional methods, which frequently rely on rigid assumptions and simple models, AI can process massive, diverse datasets over time, revealing hidden patterns that enable more accurate and personalized predictions \cite{rasmussen2006gaussian, zhang2011multi}. For example, deep learning techniques like recurrent neural networks (RNNs) excel at analyzing data gathered from multiple doctor visits, helping to refine predictions. These models improve their ability to predict how the disease will progress as more data is collected over time \cite{schuster1997bidirectional}. The increasing interest in state-space models, graph-based methods, and digital twin frameworks is even more intriguing. These new approaches are focusing on more individualized systems that replicate how Alzheimer's might progress in a particular patient rather than using one-size-fits-all models. This allows clinicians to offer more customized therapy options \cite{friston2003dynamic, kipf2017semi}. These technologies are advancing the prediction and treatment of Alzheimer's disease and bringing us one step closer to completely personalized medicine by fusing neuroscience, artificial intelligence, and advanced systems modeling \cite{zhang2011multi}.

\subsection{AI-Powered Modeling and Multimodal Integration in Alzheimer’s Disease Research}

Alzheimer’s disease (AD), the most common form of dementia, is a progressive neurodegenerative condition that significantly impairs memory, cognitive function, and daily independence over time \cite{nagarajan2025comprehensive}. Its insidious and often asymptomatic onset poses major challenges to early detection, thereby delaying timely intervention and personalized treatment planning \cite{kale2024ai}. Addressing this diagnostic complexity has become a critical research imperative, especially as the global burden of AD continues to escalate.

In response, advancements in AI—particularly in ML and DL—have ushered in a transformative era in AD research. These technologies now play a central role in enhancing early diagnosis, refining disease subtyping, and improving individualized prognostic assessments \cite{ali2025artificial}. By leveraging large-scale, multimodal datasets—encompassing neuroimaging, clinical assessments, cognitive evaluations, and molecular profiling—AI systems have facilitated the creation of high-performance diagnostic and predictive models \cite{iqbal2024progress}.

As AI methodologies grow increasingly sophisticated, their integration into both research and clinical practice has deepened, offering novel pathways for precision medicine. Among these innovations, the digital twin (DT) paradigm—defined as a dynamic, data-driven virtual representation of an individual—has emerged as a promising framework for personalizing disease forecasting and optimizing therapeutic strategies \cite{ashraf2024digital}. Together, these developments underscore the vital role of AI in reimagining the future of Alzheimer’s disease management.

\subsection{Neuroimaging, Deep Learning, and Ensemble Methods}

Recent advances in AI, particularly in DL, have revolutionized the landscape of early AD detection. By enabling automated feature extraction from high-dimensional, heterogeneous data such as structural magnetic resonance imaging (sMRI), positron emission tomography (PET \textit{-- e.g., FDG-PET for metabolism or Amyloid/Tau PET for pathology}), genetic information (\textit{e.g., APOE genotype}), and cognitive assessments, DL-based methods have consistently outperformed traditional diagnostic approaches in sensitivity and accuracy \cite{nagarajan2025comprehensive}. One of the most actively explored domains in AI-assisted AD research is the application of convolutional neural networks (CNNs) to neuroimaging data, \textit{particularly sMRI and PET}. CNNs excel in capturing spatial hierarchies and extracting complex patterns inherent in imaging modalities. Studies by \cite{kim2022}, \cite{bae2020}, and \cite{liu2022} have successfully employed CNNs to distinguish between cognitively normal (CN), mild cognitive impairment (MCI), and AD subjects using \textit{structural MRI scans derived from datasets like ADNI}. These models reported high classification accuracies and showcased the clinical potential of CNNs in early diagnosis and risk stratification.

Building upon foundational CNN architectures, recent work has sought to incorporate domain-specific insights into model design. \cite{odimayo2024} introduced SNeurodCNN, a structurally-informed CNN that integrates prior knowledge about neurodegenerative patterns in AD. \cite{ElAssy2024} developed a dual-branch CNN optimized for both early and late-stage AD detection, achieving over 99\% classification accuracy. In addition to architecture modifications, ensemble methods and hybrid spatial representations have further elevated diagnostic robustness. \cite{rahman2025} and \cite{islam2018brain} demonstrated that 2D and 3D CNNs, when combined, offer enhanced sensitivity to subtle structural anomalies. Complementary studies by \cite{mahanty2024effective} using an enhanced Xception framework and \cite{hazarika2022experimental} with modified DenseNet architectures underline the value of model diversity and regularization in mitigating overfitting and improving generalizability.

\subsection{Multimodal Integration and Diagnostic Synergy}

Multimodal deep learning, which integrates data from diverse sources, has emerged as a particularly powerful approach in AD research. \cite{Kale2024} demonstrated that combining imaging, cognitive testing, and demographic information within a unified framework enhances both diagnostic precision and prognostic modeling. Similarly, \cite{lee2019predicting} and \cite{garcia2024predicting} found that integrating imaging, cerebrospinal fluid biomarkers, and genetic data improves predictions of conversion from mild cognitive impairment (MCI) to AD. In line with these developments, \cite{Ali2025} introduced a personalized AI framework capable of tracking cognitive decline over time and forecasting individualized disease trajectories—an essential step toward precision neurology.

\begin{figure}[htbp]
    \centering
    \includegraphics[scale=0.5]{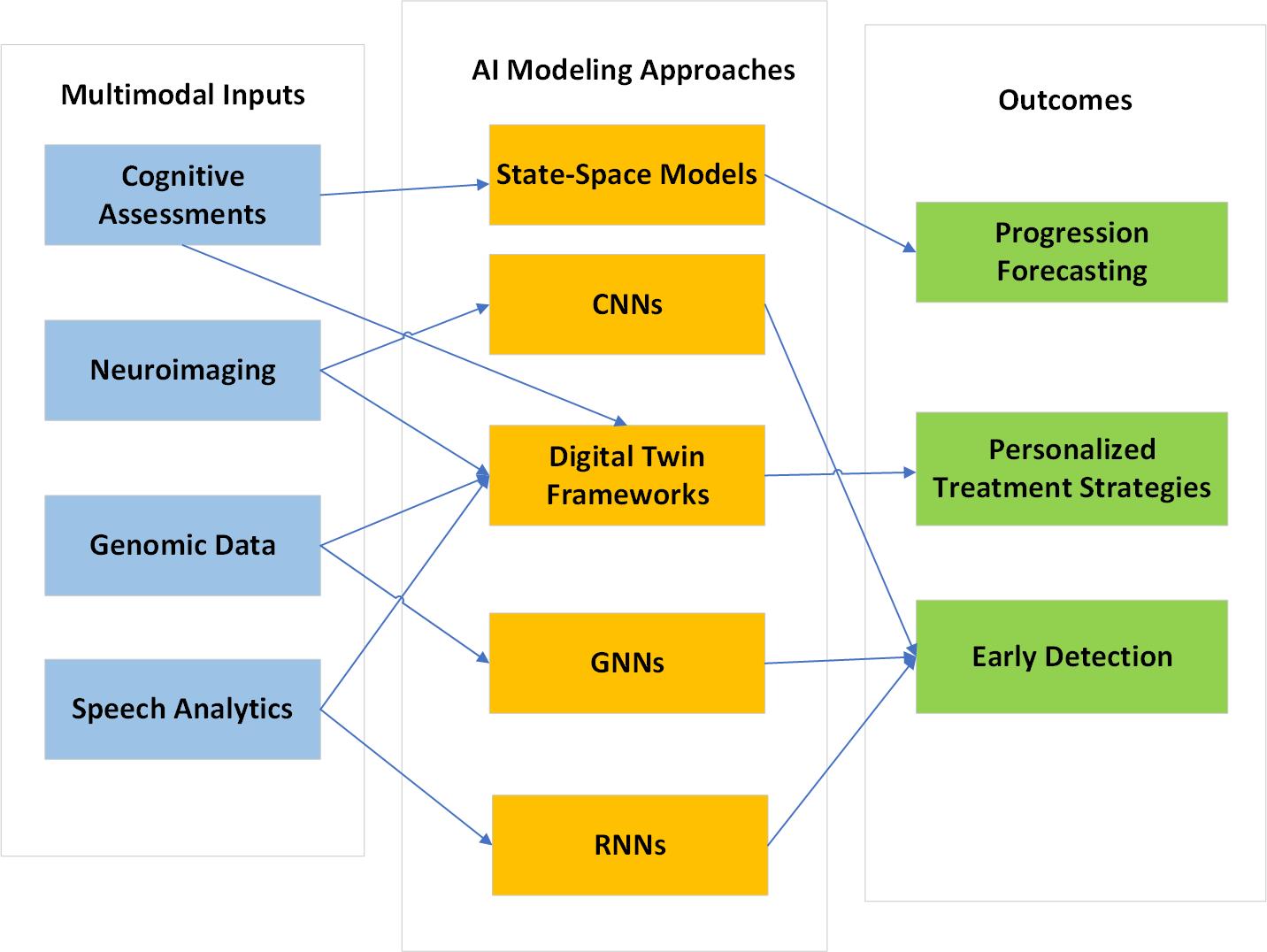} 
    \caption{Layered architecture illustrating the flow from multimodal data inputs to clinical outcomes via AI modeling approaches. Various inputs (Cognitive Assessments, Neuroimaging, Genomic Data, Speech Analytics) can be processed by different AI models (State-Space Models, CNNs, Digital Twin Frameworks, GNNs, RNNs), tailored to yield specific outcomes like Progression Forecasting, Personalized Treatment Strategies, and Early Detection. Arrows suggest typical, though not exclusive, data-model-outcome mappings.  This figure emphasizes the modularity and adaptability of AI systems for personalized AD prediction, showing how different data types and modeling strategies can be orchestrated to achieve specific clinical objectives.}
    \label{fig:Figure 9}
\end{figure}

Given the multifactorial nature of AD—which involves neuroimaging, genetics, clinical measures, and cognitive decline—multimodal data integration is critical. \cite{odusami2023explainable}, \cite{venugopalan2021}, and \cite{qiu2022} employed deep learning architectures that combine structural and functional neuroimaging with demographic and genetic data, significantly improving classification robustness and trajectory modeling. \cite{nguyen2020predicting} further advanced interpretability by integrating cognitive scores with an ensemble of 3D-ResNet and XGBoost classifiers; the inclusion of demographic data as supplementary features enhanced both accuracy and transparency. Supporting this, \cite{mahendran2022deep} and \cite{lee2024} demonstrated that incorporating longitudinal biomarkers within multimodal pipelines significantly boosts prediction performance for MCI-to-AD conversion. Figure~\ref{fig:Figure 9} illustrates the layered architecture of personalized AD progression modeling, demonstrating how multimodal data inputs are processed through specialized AI models to generate tailored clinical predictions.

\subsection{Longitudinal Progression and Predictive Analytics}

Beyond diagnostic classification, DL models have increasingly been used to capture the complex, nonlinear dynamics of AD progression over time. Temporal modeling is essential for understanding disease evolution, supporting early intervention, and enabling individualized forecasts. RNNs and their variants, such as LSTM networks, are particularly well-suited to this task due to their ability to model time-dependent patterns in longitudinal patient data.

\cite{nguyen2020predicting} proposed a deep RNN that leverages each patient’s historical clinical data to predict future cognitive status. This model outperformed simpler approaches such as Markov chains by learning rich latent representations of disease states and achieved high predictive accuracy on the Alzheimer’s Disease Neuroimaging Initiative (ADNI) dataset when forecasting cognitive trajectories over a three-year span \cite{garcia2024predicting}. Similarly, \cite{liang2021rethinking} developed a multi-task RNN model capable of simultaneously predicting cognitive scores, diagnostic labels, and ventricular volume changes across multiple timepoints. Their multi-output strategy enabled a more comprehensive view of disease progression, offering valuable insights for clinical decision-making.

Building on this work, \cite{saleh2023computer} employed ensemble LSTM networks to model sequential dependencies in multivariate neuropsychological data. These models further improved performance, especially in detecting subtle transitions from mild cognitive impairment (MCI) to AD—an area where traditional classifiers often fall short. Collectively, these studies underscore the value of deep learning architectures tailored for sequence data, reinforcing the role of dynamic modeling in both early-stage diagnosis and long-term prognosis.

In parallel, generative modeling techniques such as Conditional Restricted Boltzmann Machines (CRBMs) have emerged as powerful tools for simulating hypothetical disease trajectories. \cite{bertolini2020modeling, bertolini2021forecasting} and \cite{Fisher2019} applied CRBMs to generate in silico forecasts of disease progression under varying clinical scenarios. These models offer a sandbox environment for exploring potential intervention outcomes, ultimately supporting the development of adaptive clinical trials and personalized treatment strategies.

\subsection{State-Space and Deep Temporal Models}

Modeling the longitudinal progression of AD is crucial for forecasting individual disease trajectories and evaluating therapeutic interventions. State-space models (SSMs) provide a robust statistical framework for this purpose, representing disease progression as a latent, unobserved state that evolves over time while linking it to observed measurements, such as \textit{longitudinal cognitive scores (MMSE, ADAS-Cog), CSF biomarker levels (A$\beta$42, p-tau), or imaging-derived metrics (hippocampal volume from sMRI)}. These models, along with multistate frameworks such as Markov models, have a longstanding role... particularly in analyzing transitions between clinical stages \textit{(often defined using criteria like CDR scores)} such as normal cognition, mild cognitive impairment (MCI), and AD.

Chua and Tripodis \cite{chua2022state} exemplified the utility of SSMs by applying an adjusted local linear trend model to longitudinal neuropsychological test data, effectively capturing individual variability in cognitive decline. These models are particularly adept at handling measurement noise and accounting for within-subject correlations, thereby producing smoother and more accurate disease progression estimates. Extending this line of research, He \cite{he_framework_2024} applied state-space techniques to neural oscillatory dynamics captured via EEG, revealing sleep-related brainwave patterns as potential early and non-invasive indicators of neurodegeneration.

Building on traditional SSM approaches, recent innovations have integrated deep learning techniques to better capture the high-dimensional and nonlinear nature of disease progression. Burkhart et al. \cite{burkhart2024unsupervised}; for instance, proposed a mixture-based state-space framework for unsupervised multimodal trajectory modeling, successfully stratifying individuals by risk levels based on their latent disease pathways. These dynamic modeling strategies underscore the growing value of temporal models in achieving more granular and predictive insights into AD evolution.

Complementing SSMs, deep learning-based temporal models have gained prominence for longitudinal prediction. Liang et al. \cite{liang2021rethinking} developed a multi-task recurrent neural network (RNN) model capable of simultaneously forecasting cognitive scores, diagnostic outcomes, and ventricular volume changes over time. This multi-output learning strategy offers a holistic view of disease progression and supports early clinical decision-making. Saleh et al.\cite{saleh2023computer} further advanced temporal modeling by employing ensemble long short-term memory (LSTM) networks to capture sequential dependencies within multivariate neuropsychological data. These dynamic approaches significantly outperform static classifiers, especially in detecting the subtle progression from MCI to AD.
Additionally, generative models such as CRBMs have emerged as valuable tools for simulating hypothetical disease trajectories. Bertolini et al. \cite{bertolini2020modeling, bertolini2021forecasting} and Fisher et al. \cite{Fisher2019} demonstrated the potential of CRBMs to model alternative disease scenarios and evaluate intervention outcomes in silico, thereby aiding the design of adaptive clinical trials and supporting personalized treatment planning.

\subsection{Network Dynamics for Alzheimer's Prediction}

Understanding the dynamic organization of brain networks is critical for improving early diagnosis and progression modeling in AD. Zhang et al. \cite{zhang2023dimension} found that frequency-specific coactivation patterns in brain networks correlate with different AD stages, suggesting that dynamic brain connectivity metrics can serve as useful biomarkers. These neurophysiological markers are particularly valuable for inclusion in state-space and deep learning frameworks, where real-time, high-resolution data enable precise modeling of disease evolution. As computational methods continue to advance, integrating these functional brain measurements into multimodal pipelines will be key for enhancing early AD diagnostics.

In addition to signal-level modeling, researchers have increasingly focused on brain network synchronization and connectivity dynamics. Jalili \cite{jalili2015multivariate} reviewed various methods for multivariate synchronization analysis in EEG data, highlighting their utility in characterizing interactions across distributed brain regions. These approaches support the development of dynamic models that reflect real-time neural coordination, which is often disrupted in early AD. Building on this, Padole et al. \cite{padole2020characterization, padole2022early} proposed graph-based representations of evolving brain states and applied these to the prediction of AD progression. Their work combined graph theory and SSMs to model time-evolving functional networks, enabling better characterization of cognitive transitions. On a broader scale, population-level modeling efforts like those of Yashin et al.\cite{yashin2016analysis} have investigated dementia incidence trends over decades using longitudinal survival models. These studies collectively suggest that integrating network dynamics into personalized models offers new insights into the multiscale nature of AD—spanning from individual brain oscillations to population epidemiology.

Expanding upon the concept of heterogeneity in disease progression, Chang et al.\cite{Chang2024} and Wen et al.\cite{wen2024genetic}  further refined predictive analytics through patient stratification. Chang’s deep clustering model categorized AD patients into demographically distinct groups based on age, race, and sex. Wen’s use of Surreal-GAN for unsupervised representation learning identified neuroanatomical patterns aligned with APOE status and other genomic markers. These contributions reveal underlying heterogeneity in AD pathology and open doors to stratified care models. Similarly, Kim et al. \cite{Kim2020} and van der Haar et al. \cite{van2023} highlighted the importance of anatomical and phenotypic subtypes. By identifying focal cortical atrophy and progression-based clusters, these studies demonstrated how disease manifestation varies widely and requires nuanced diagnostic tools.

\subsection{Language, Cognition, and Biomarker Discovery Beyond Imaging}

Neurophysiological signals—especially EEG and fMRI—remain critical in the early detection and modeling of AD due to their ability to capture functional brain dynamics in real time. While structural imaging techniques like MRI and PET are indispensable for identifying anatomical and molecular abnormalities, EEG and fMRI provide insight into the brain's temporal and spatial activity patterns. Wan et al. \cite{wan2024beyond} introduced a complementary-ensemble methodology for detecting microstate fluctuations in EEG data, which may be indicative of early cognitive decline. Similarly, EEG-based approaches by Fouad and Labib  \cite{fouad2023identification} revealed high predictive accuracy using deep residual networks, demonstrating EEG’s promise as a low-cost, accessible diagnostic modality. Shan et al. \cite{Shan2021} refined amyloid positivity prediction using domain-level cognitive scores, offering a methodology potentially more sensitive to early impairments than traditional global scores.

Beyond imaging, the inclusion of non-imaging data—especially language and cognitive markers—further expands AI’s diagnostic scope. Mao et al.  \cite{Mao2023} developed AD-BERT, a transformer model leveraging clinical narratives to detect signs of cognitive decline, while Asgari et al.  \cite{Asgari2017} extracted linguistic features from spontaneous speech to classify mild cognitive impairment (MCI), illustrating the power of subtle behavioral signals in early AD detection. In parallel, in the molecular domain, Shigemizu et al.  \cite{shigemizu2023classification} and Alamro et al.  \cite{alamro2023exploiting} employed feature selection and gene interaction networks to identify AD-associated genes and potential therapeutic targets, bridging genotype with phenotype and aligning with imaging findings for a more comprehensive diagnostic approach.

\subsection{Validation, Tau Imaging, and Histopathological Alignment}

Validating AI predictions against pathological ground truth is vital for clinical adoption. Signaevsky et al. \cite{Signaevsky2019} and Ushizima et al.\cite{ushizima2022deep} developed deep learning models to quantify tau tangles and amyloid plaque burden from histopathological slides, enhancing biomarker quantification and providing critical benchmarks for validating imaging-based models. Building on this, Park et al.\cite{Park2023} integrated CNN and LSTM architectures with tau PET data to improve stage classification. Similarly, Massa et al. \cite{massa2022added} demonstrated that pairing semiquantitative FDG-PET analysis with machine learning models improves differential diagnosis between MCI due to AD and MCI due to Lewy body disease, highlighting the importance of refined radiomic features in clinical differentiation.

\subsection {Digital Twins, Generative Modeling, and Personalization}

The convergence of AI, temporal modeling, and multimodal data integration (\textit{combining data streams like imaging, biomarkers, clinical scores, and genetics}) has catalyzed the development of DT technologies in healthcare. A digital twin is a continuously updated, virtual model that mirrors an individual's biological state and disease progression based on their accumulating data. Originally developed in engineering, this concept is now transforming chronic disease management, including AD, by shifting clinical paradigms from population-level inference to individualized, data-driven decision support  \cite{singh2022applications,sprint2024building,ashraf2024digital}.

\begin{figure}[htbp]
    \centering
    \includegraphics[scale=0.5]{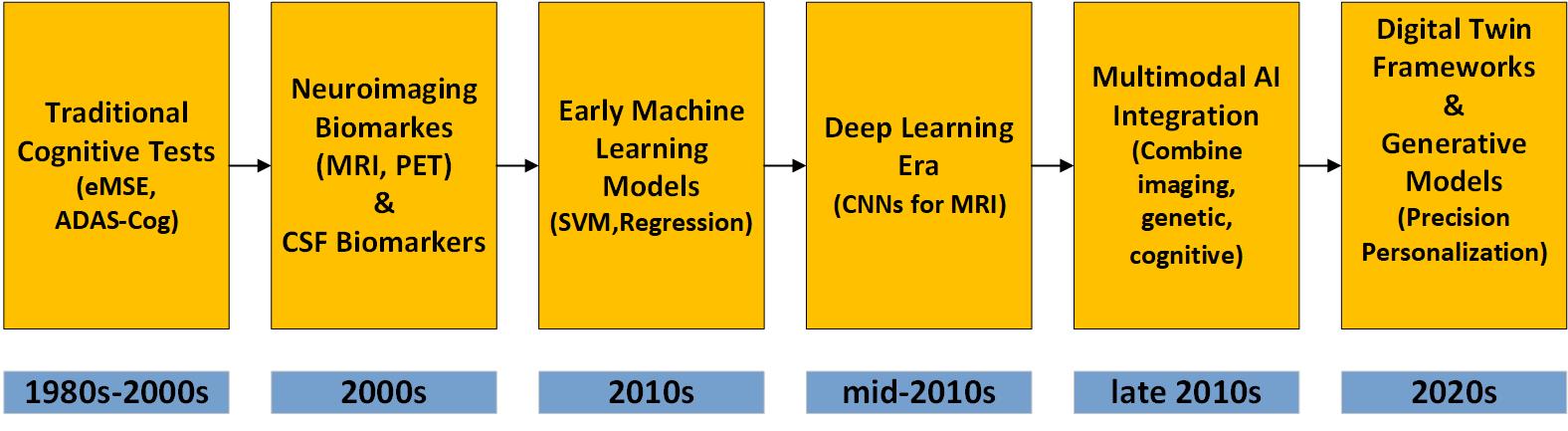} 
    \caption{Timeline depicting the historical evolution of methodologies used in AD research and diagnostics, spanning from the 1980s to the 2020s. It traces the progression from traditional cognitive testing, through the introduction of neuroimaging and CSF biomarkers, the application of early machine learning models, the rise of deep learning, the integration of multimodal AI, and the current emergence of sophisticated digital twin frameworks aimed at precision personalization. This timeline illustrates the rapid advancement and increasing computational sophistication of tools applied to AD research, highlighting the accelerating trend towards data-driven, personalized approaches leveraging AI.}
    \label{fig:Figure 10}
\end{figure}

In this evolving framework, deep generative models have emerged as powerful tools for simulating disease trajectories. The broader evolution of Alzheimer’s disease modeling, from traditional cognitive testing to AI-driven digital twin frameworks, is summarized in Figure ~\ref{fig:Figure 10}, providing historical context for recent advances in personalized simulation approaches. Fisher et al. \cite{Fisher2019} pioneered the application of a Conditional Restricted Boltzmann Machine (CRBM) to generate personalized forecasts of clinical and biomarker outcomes in AD. Building on this foundation, Bertolini et al.  \cite{bertolini2021forecasting} utilized generative modeling techniques to create digital twins based on real-world baseline data, enabling individualized projections and improving clinical trial efficiency through the simulation of control groups.

Recent advances have expanded the digital twin paradigm across biological scales and data modalities. Sprint et al.  \cite{sprint2024building} introduced the Human Digital Twin (HDTwin) framework, leveraging large language models (LLMs) to integrate genomic, imaging, behavioral, and biosensor data into dynamic patient profiles that evolve with real-time inputs. Complementing this macroscale approach, Ren et al.  \cite{ren2025single} proposed a single-cell digital twin system that models therapeutic responses at the cellular level to identify drug targets in AD. Together, these innovations demonstrate how digital twins can operate from molecular mechanisms to whole-patient simulations, offering new opportunities for in silico treatment testing, trial enrichment, and proactive clinical management.

As real-world health data become increasingly multimodal and longitudinal, digital twin systems are positioned to become a cornerstone of precision neurology. By enabling personalized forecasts, simulating disease progression under various scenarios, and guiding targeted interventions, digital twins reduce clinical uncertainty and promote more ethical, individualized decision-making. Supported by generative models like CRBMs and advanced AI architectures, digital twin technologies represent a transformative shift toward patient-specific disease modeling and are poised to significantly advance both therapeutic development and clinical care in Alzheimer's disease  \cite{singh2022applications,sprint2024building,ashraf2024digital,ren2025single}.

\subsection{Interpretability, Generalizability, and Clinical Decision Support}

Collectively, recent innovations represent a paradigm shift from traditional statistical analyses toward AI-driven, individualized modeling of AD. The fusion of multimodal data, state-space modeling, deep learning, and generative simulation has laid the groundwork for constructing rich, adaptive models of cognitive decline. These approaches enable early identification of disease onset and progression at a level of granularity and personalization not possible with conventional methods. By capturing nonlinear trajectories, integrating latent variables, and modeling uncertainty, such frameworks provide a more nuanced and responsive understanding of neurodegeneration. In parallel, they offer tools to optimize trial design, patient stratification, and treatment decision-making, promising to accelerate translational progress in AD research and care.

Despite these advancements, challenges remain for clinical integration, particularly regarding model interpretability. Chang et al. \cite{Chang2024}, Nguyen et al.\cite{nguyen2022ensemble}, and Liang et al.\cite{liang2021rethinking} addressed this issue by incorporating SHAP values, saliency maps, and multi-task learning weights to elucidate model decisions—efforts that are essential for fostering clinician trust and meeting regulatory standards. Additionally, ensuring model generalizability is critical. Wang et al. \cite{wang2022proof} proposed hybrid training frameworks that maintain performance even with limited datasets, while Lee et al. \cite{lee2024multimodal} developed multimodal predictive models that demonstrated robust results across independent cohorts. These studies underscore the importance of data diversity, model regularization, and external validation for real-world applicability.

Finally, broader perspectives from Ali and Manda \cite{ali2025artificial} and Kale et al. \cite{kale2024ai} emphasize the transformative role of AI in clinical workflows. Their overviews link diagnostic modeling, prognostic forecasting, and clinical decision support, positioning AI as a comprehensive and integrative tool for advancing precision neurology in Alzheimer's disease.

In summary, the convergence of dynamic modeling, artificial intelligence, and multimodal integration is redefining the future of Alzheimer’s disease prediction and care. These technologies enable earlier, more accurate, and individualized prognostication, supporting timely interventions that could improve patient outcomes and reduce societal burden. The reviewed studies collectively demonstrate how AI, machine learning, and deep learning approaches are transforming the understanding, diagnosis, and treatment of AD by integrating neuroimaging, cognitive assessments, genomics, and speech analytics into interpretable, scalable models. Innovations in digital twins, transformer-based language models, and multi-task architectures continue to push the boundaries of AD research, paving the way for precision medicine and population health strategies alike.

\section{Data Resources and Challenges in AI-Driven Personalized Modeling of Alzheimer's Disease Progression}

\subsection{Data Resources for AD Progression}

Accurate and personalized prediction of AD progression relies heavily on the availability of large, longitudinal, and multimodal datasets. These datasets provide the necessary clinical, imaging, cognitive, and biomarker information for training and validating advanced AI models. In this section, we summarize the major publicly available and referenced data resources used by the studies reviewed, highlighting their characteristics and role in advancing personalized AD forecasting.

\subsection{Common Data Modalities in AD Progression Research} 
The studies reviewed leverage a variety of data types commonly collected in large observational cohorts, particularly the ADNI study. Understanding these modalities is crucial for appreciating the inputs to the AI models discussed. Key data types include:
\begin{itemize}
    \item \textbf{Neuroimaging:} Structural Magnetic Resonance Imaging (sMRI), primarily T1-weighted scans yielding measures like regional brain volumes (e.g., hippocampus) and cortical thickness; Positron Emission Tomography (PET), including [18F]fluorodeoxyglucose (FDG-PET) measuring cerebral metabolic rate, Amyloid-PET (e.g., using [18F]florbetapir, AV45) quantifying amyloid plaque burden, and Tau-PET (e.g., using [18F]flortaucipir, AV1451) assessing neurofibrillary tangle distribution. Diffusion Tensor Imaging (DTI) and resting-state functional MRI (rs-fMRI) are also used, often for deriving brain connectivity measures.
    \item \textbf{Fluid Biomarkers:} Measurements obtained typically from Cerebrospinal Fluid (CSF), including Amyloid-beta 1-42 (A$\beta$42), total Tau (t-tau), and phosphorylated Tau at threonine 181 (p-tau181). Plasma biomarkers (e.g., p-tau variants) are emerging but less represented in the core longitudinal studies reviewed here.
    \item \textbf{Genetic Information:} Most prominently, the Apolipoprotein E (APOE) genotype, focusing on the presence and count of the risk-associated $\epsilon$4 allele. Some studies incorporate wider genomic data like single nucleotide polymorphisms (SNPs).
    \item \textbf{Clinical and Cognitive Assessments:} Longitudinal scores from a battery of standardized tests, including global cognition measures like the Mini-Mental State Examination (MMSE) and Alzheimer's Disease Assessment Scale-Cognitive subscale (ADAS-Cog), functional measures like the Functional Assessment Questionnaire (FAQ), and staging measures like the Clinical Dementia Rating sum of boxes (CDR-SB).
    \item \textbf{Demographic Data:} Basic information such as age at assessment, sex, and years of education.
\end{itemize}
The longitudinal nature of these datasets, with repeated measurements over multiple years, is fundamental for training models capable of predicting temporal progression patterns.

\subsubsection{Alzheimer's Disease Neuroimaging Initiative (ADNI)}

The ADNI is the most widely utilized dataset in personalized AD research. It provides longitudinal magnetic resonance imaging, PET, cognitive assessments, cerebrospinal fluid biomarkers, and clinical data across the aging and dementia spectrum. Many studies reviewed in this paper rely on ADNI to develop and benchmark predictive models. Nguyen et al. \cite{nguyen2020predicting} employed ADNI to train deep recurrent neural networks capable of forecasting cognitive scores over time. García-Gutiérrez et al. \cite{garcia2024predicting} applied multitask deep learning to predict longitudinal changes in brain metabolism and conversion from MCI to dementia using ADNI data. Liang et al.\cite{liang2021rethinking} designed a multitask RNN model using ADNI to simultaneously forecast cognitive scores, diagnosis, and anatomical changes. Bertolini et al. \cite{bertolini2020modeling,bertolini2021forecasting} developed digital twins using Conditional Restricted Boltzmann Machines, trained entirely on ADNI baseline data. A wide range of deep learning architectures were also trained and tested on ADNI MRI and cognitive data, including Mahanty et al. \cite{mahanty2024effective}, Islam et al. \cite{islam2018brain}, Kim et al. \cite{kim2022deep}, Odimayo et al.\cite{odimayo2024structure}, El-Assy et al. \cite{ElAssy2024}, and Rahman et al.\cite{rahman2025alzheimer}. In the domain of explainable and generative modeling, Wen et al. \cite{wen2024genetic} and Chang et al. \cite{Chang2024} used ADNI neuroimaging and genetic data to explore latent subtypes and APOE-related variation. Odusami et al. \cite{odusami2023explainable} further used ADNI to develop explainable convolutional models using saliency-based interpretations.

\subsubsection{TADPOLE Challenge Dataset}

Derived from the ADNI database, the TADPOLE dataset focuses on forecasting future clinical diagnosis, cognitive trajectories, and neuroimaging biomarker evolution. It was specifically created for a global modeling challenge and has since become a benchmark for disease progression forecasting. Burkhart et al. \cite{burkhart2024unsupervised} built unsupervised multimodal trajectory models using TADPOLE-derived features, enabling stratification of individuals based on latent risk trajectories. Nguyen et al.\cite{nguyen2020predicting} also evaluated their longitudinal deep learning framework on TADPOLE-defined forecasting targets, illustrating the dataset’s role in modeling future disease states.

\subsubsection{National Alzheimer’s Coordinating Center (NACC)}

The NACC database aggregates longitudinal clinical, cognitive, and neuropathological data from Alzheimer’s Disease Centers across the United States. Its diversity makes it particularly useful for external validation of models trained on ADNI. Liu et al. \cite{liu2022generalizable} evaluated their CNN-based diagnostic model using MRI data from NACC participants to assess generalizability. Qiu et al.\cite{qiu2022multimodal} developed a multimodal ensemble model incorporating both ADNI and NACC data. Additionally, Mahendran et al. \cite{mahendran2022deep} highlighted the importance of using validation cohorts similar to NACC to demonstrate the robustness of longitudinal predictive pipelines.

\subsubsection{Custom and Synthetic Datasets (EEG, Omics, Simulated Cohorts)}

\begin{figure}[htbp]
    \centering
    \includegraphics[scale=0.5]{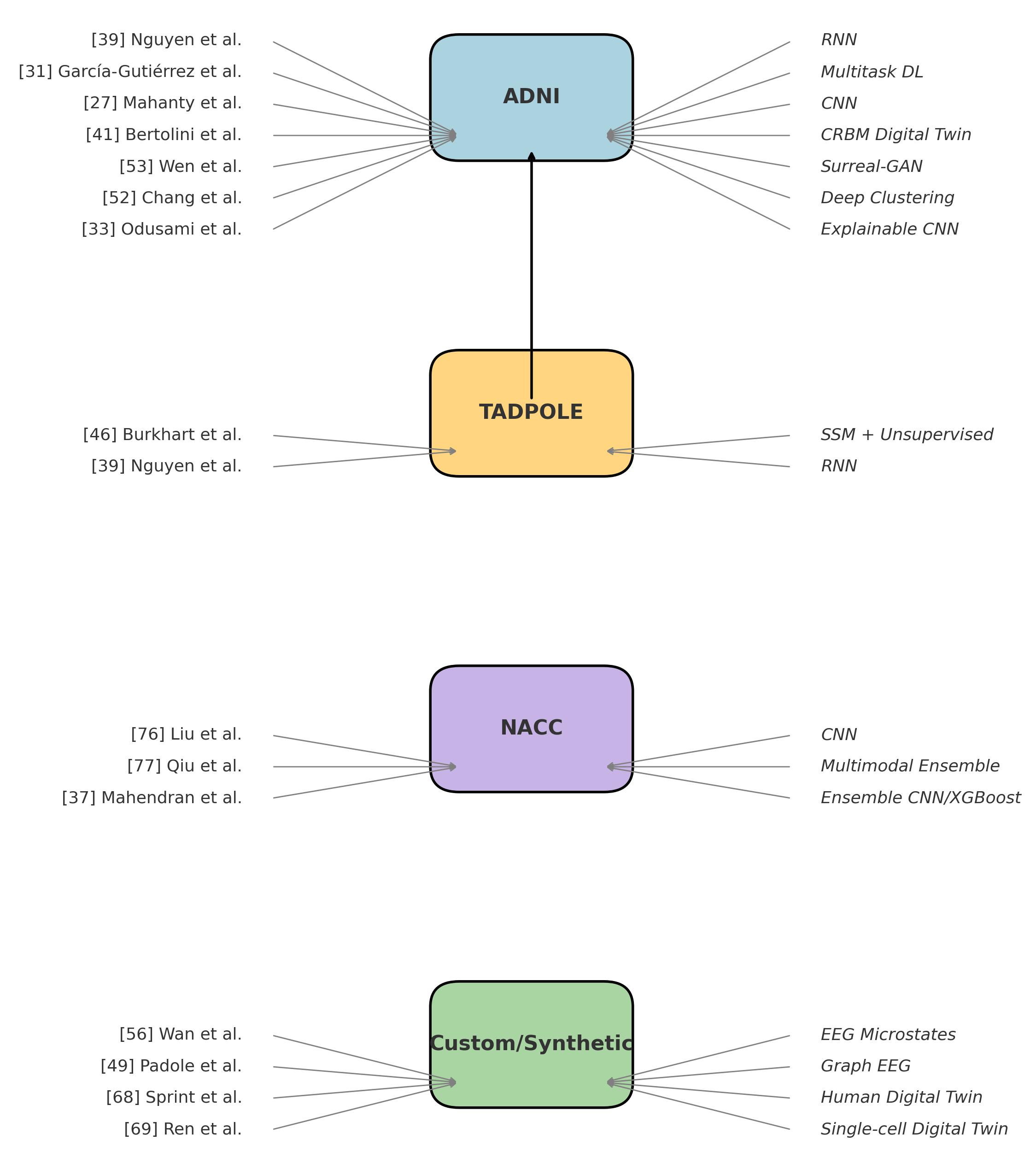} 
    \caption{Diagram mapping key datasets used in personalized AD progression research to specific studies reviewed in this paper (references indicated by numbers). It highlights the central role of ADNI, the derivative TADPOLE challenge dataset, the use of NACC often for validation, and the application of custom or synthetic datasets (e.g., EEG, omics) for specialized or novel methodological investigations. This mapping visualizes the data ecosystem supporting AI research in AD progression, emphasizing the reliance on major public datasets like ADNI while showcasing the diversification towards complementary resources for validation and innovation.}
    \label{fig:Figure 11}
\end{figure}

Several studies advanced novel modeling techniques using EEG-derived, omics-based, or synthetic clinical datasets. Wan et al.\cite{wan2024beyond} used custom-collected EEG data to detect microstate fluctuations associated with early cognitive decline. Padole et al.\cite{padole2020characterization,padole2022early} proposed graph-based representations of evolving brain states based on functional EEG data. In parallel, Sprint et al. \cite{sprint2024building} created Human Digital Twins (HDTwins) using synthetic multimodal data sources designed to reflect real-world patient complexity. Ren et al. \cite{ren2025single} introduced a single-cell digital twin framework leveraging omics datasets to identify therapeutic targets and simulate cellular disease responses. The relationships between datasets and the studies that utilized them are summarized in Figure~\ref{fig:Figure 11}.

This mapping highlights the central role of ADNI and the increasing use of auxiliary datasets to enhance model robustness and generalization. Together, these datasets form the empirical foundation for AI-driven personalized AD modeling. ADNI remains the primary resource due to its breadth and longitudinal depth, while TADPOLE and NACC enhance evaluation and generalization. Meanwhile, custom EEG and biological datasets are supporting innovation in underexplored areas such as brain network dynamics and cellular modeling. The synergy between these diverse data resources and advanced AI methodologies is accelerating progress toward precision neurology in AD.

\subsection{Data Challenges in Personalized Modeling}
\label{sec:data_challenges}

The successful realization of accurate and reliable personalized prediction models for AD progression is intrinsically linked to the characteristics of the available data. While large-scale, longitudinal initiatives such as the ADNI have provided the community with invaluable datasets, researchers consistently grapple with significant inherent challenges associated with collecting and utilizing clinical and biological data over extended periods in the context of complex, slowly progressing neurodegenerative disorders. Overcoming these data hurdles is not merely a technical prerequisite but a fundamental necessity for constructing AI models that can genuinely capture the nuances of individual patient trajectories and offer clinically meaningful prognostic insights. The transition from population-level averages to individualized forecasts demands methodologies capable of handling the imperfections and complexities pervasive in real-world patient data.

Several formidable data challenges commonly impede the development of robust personalized AD progression models. A primary issue is \textbf{high dimensionality}. Patient datasets frequently amalgamate a vast array of features derived from heterogeneous sources. This includes potentially millions of features from high-resolution neuroimaging modalities like structural sMRI or PET, thousands of genetic markers such as single nucleotide polymorphisms (SNPs), comprehensive batteries of cognitive and functional assessments yielding numerous scores, and detailed demographic and clinical history variables. This extreme dimensionality, often referred to as the "curse of dimensionality," not only escalates computational demands significantly but also substantially increases the risk of model overfitting, where algorithms learn spurious correlations or noise specific to the training sample rather than generalizable underlying biological patterns relevant to disease progression. Effectively identifying the truly informative features within this high-dimensional space remains a critical challenge.

Furthermore, the longitudinal nature essential for tracking disease progression introduces challenges related to \textbf{data sparsity and irregular sampling intervals}. Clinical assessments and biomarker measurements are typically conducted at non-uniform time points, often dictated by clinical practice guidelines, research protocol constraints, or patient availability, rather than modeling convenience. Moreover, the duration of follow-up and the total number of visits can vary dramatically across participants within a cohort. This results in time-series data that is sparse for many individuals, containing large gaps between observations. Standard time-series models often assume regular sampling, and adapting them or developing new models to handle such irregularity without introducing bias is complex. Modeling smooth, continuous trajectories from these fragmented observations requires sophisticated approaches.

Compounding these issues is the pervasive problem of \textbf{missing data}. It is exceedingly rare for longitudinal datasets in AD research to be complete. Participants may miss scheduled visits, specific tests or scans might not be performed due to technical issues or patient intolerance, data recording errors can occur, or protocols might evolve over the long duration of a study, leading to missing values for certain variables or entire modalities at specific time points. While traditional methods like mean or median imputation are simple, they fail to account for relationships between variables and can severely distort the data distribution, potentially masking true effects or introducing significant bias into downstream analyses and model predictions, particularly in complex multimodal settings. More advanced imputation techniques are required to handle missingness in a manner that preserves the integrity of the data.

Another significant hurdle arises from \textbf{imbalanced datasets}. The inherent heterogeneity of AD means that progression rates vary widely. Observational cohorts often disproportionately represent individuals with slower disease progression (\textit{e.g., remaining stable MCI for many years}) or those in earlier clinical stages (e.g., Mild Cognitive Impairment, MCI) compared to those experiencing rapid decline (\textit{e.g., converting from MCI to AD within a year}) or those already in later dementia stages. Similarly, specific etiological subtypes of AD, or individuals carrying certain genetic risk factors (like APOE $\epsilon$4 homozygotes) or protective factors, might constitute only a small fraction of the overall cohort. AI models trained on such imbalanced data naturally tend to develop a bias towards the majority class, leading to poor predictive performance for the underrepresented, yet potentially clinically critical, minority groups. This limitation hinders the development of truly personalized models that perform equitably across the full spectrum of disease presentations and progression speeds.

Finally, even within large overall consortia, the investigation of specific research questions or the development of models tailored to particular patient subgroups often faces the challenge of \textbf{small effective sample sizes}. For instance, studies focusing on early-onset AD, specific genetic mutations beyond APOE, individuals from underrepresented demographic groups, or patients with particular comorbidities might find that the number of participants meeting the precise inclusion criteria is limited. Training complex AI models, especially deep learning architectures with millions of parameters, typically requires substantial amounts of data to avoid overfitting and ensure generalizability. Small sample sizes severely constrain model complexity and make robust validation difficult, hindering the discovery of reliable patterns specific to these subgroups.

Fortunately, the field of AI provides a growing arsenal of advanced techniques designed to confront these data challenges more effectively than traditional statistical methods \cite{soykan2025}. Key AI-driven strategies often focus on either intelligently estimating or filling in missing information or, perhaps more powerfully, on augmenting the available data through the generation of realistic synthetic patient samples.

\subsubsection{Synthetic Data Generation Using Variational Autoencoders (VAEs)}

\label{subsec:vaes}
VAEs represent a prominent class of deep generative models that excel at learning complex, high-dimensional data distributions and generating new samples from them. A VAE architecture typically comprises two interconnected neural networks: an \textit{encoder} (also called the recognition network) and a \textit{decoder} (the generative network). The encoder's role is to compress an input data point (e.g., an MRI scan, a set of clinical scores) into a lower-dimensional representation within a latent space. Unlike standard autoencoders, VAEs impose a probabilistic structure on this latent space, typically modeling it as a multivariate Gaussian distribution. The encoder outputs the parameters (mean and variance) of the distribution corresponding to the input. The decoder then takes a point sampled from this latent distribution and attempts to reconstruct the original input data. The VAE is trained by simultaneously optimizing two objectives: minimizing the \textit{reconstruction error} (ensuring the generated data resembles the input) and minimizing the \textit{Kullback-Leibler (KL) divergence} between the learned latent distribution and a prior distribution (usually a standard normal distribution), which acts as a regularizer encouraging a smooth and well-structured latent space.

By sampling points from the prior distribution in the latent space and passing them through the trained decoder, VAEs can generate entirely new, synthetic data points that exhibit characteristics similar to the original training data. Within AD research, VAEs have demonstrated considerable utility. They have been successfully applied to generate realistic synthetic neuroimaging data, such as T1-weighted MRI scans or FDG-PET images, which can then be used to augment limited training datasets, thereby improving the robustness and generalization capability of subsequent diagnostic or prognostic classification models. Furthermore, VAEs can be adapted to model sequential data, enabling the generation of plausible longitudinal trajectories of cognitive scores or biomarker measurements. This synthetic longitudinal data can help fill gaps in sparse patient records or create balanced datasets containing representative examples of different progression speeds or patterns, addressing the class imbalance problem. Additionally, the learned latent space of a VAE can itself be valuable, potentially capturing underlying disease phenotypes or stratifying patients in a data-driven manner, offering insights beyond direct data generation.

\subsubsection{Synthetic Data Generation Using Generative Adversarial Networks (GANs)}
\label{subsec:gans}
GANs constitute another highly influential paradigm for generative modeling, renowned for their ability to produce remarkably realistic synthetic data, particularly images \cite{goodfellow2016}. The core concept of a GAN involves a dynamic interplay, framed as a zero-sum game, between two neural networks: a \textit{Generator} ($G$) and a \textit{Discriminator} ($D$). The Generator's objective is to learn the distribution of the real data and create synthetic samples (e.g., synthetic brain scans) from a random noise input. The Discriminator's task is to evaluate inputs and determine whether they are real samples from the original dataset or synthetic ('fake') samples produced by the Generator. These two networks are trained concurrently: $D$ is trained to maximize its accuracy in distinguishing real from fake, while $G$ is trained to minimize the probability that $D$ correctly identifies its output as fake, effectively trying to 'fool' the Discriminator. This adversarial process drives the Generator to produce increasingly plausible data that closely mimics the statistical properties of the real dataset. Convergence is reached, ideally, at a Nash equilibrium where the Generator produces data indistinguishable from real data, and the Discriminator is unable to differentiate better than random chance.

In the realm of AD research, GANs have been explored for several applications, often complementing or competing with VAEs. They are particularly noted for generating high-fidelity synthetic neuroimaging data, with studies suggesting that GAN-generated MRIs or PET scans can sometimes achieve higher visual realism or preserve finer anatomical details compared to those from VAEs. This capability is crucial for augmenting training datasets used for developing image-based diagnostic classifiers or segmentation algorithms. Specialized GAN architectures, such as Conditional GANs (cGANs), allow for controlled data generation by providing auxiliary information (e.g., clinical diagnosis, age) as input to both the Generator and Discriminator. This enables the generation of synthetic data corresponding to specific patient subgroups or disease stages, directly addressing class imbalance issues by synthesizing samples for underrepresented categories. Furthermore, extensions like Recurrent GANs or Time-series GANs are being developed to generate realistic longitudinal sequences of multimodal data, potentially capturing the complex temporal dynamics of AD progression. Despite their power, training GANs can be notoriously unstable, requiring careful tuning of hyperparameters and network architectures. Critically, as with VAEs, rigorous validation is essential to ensure that GAN-generated synthetic data accurately reflects the true data distribution and does not introduce unintended artifacts or biases that could compromise the reliability of models trained on augmented datasets.

\subsubsection{Other AI-Driven Mitigation Strategies}

\label{subsec:other_ai_strategies}
While synthetic data generation offers a powerful means to augment and balance datasets, other AI methodologies also play a significant role in mitigating data challenges in personalized AD modeling.

Advanced techniques for handling \textbf{missing data imputation} have emerged from the machine learning community. AI-based methods often outperform traditional statistical imputation (like mean, median, or regression imputation) by leveraging the complex, non-linear relationships present in high-dimensional, multimodal data. For instance, autoencoder-based imputation involves training an autoencoder on the available data and then using the trained model to reconstruct the missing values based on the observed features for a given patient. Alternatively, algorithms like k-Nearest Neighbors (k-NN) imputation estimate missing values based on the values observed in the most similar patients (neighbors) in the feature space. Methods like MissForest, based on random forests, iteratively predict missing values for each variable using the others. These approaches can capture intricate multivariate dependencies, leading to more accurate and plausible imputations compared to simpler univariate methods, thereby preserving more information for downstream modeling.

Another highly relevant strategy, particularly when dealing with limited sample sizes for specific tasks or subgroups, is \textbf{Transfer Learning}. This paradigm leverages knowledge gained from solving one problem (the source task, often trained on a large dataset) and applies it to a different but related problem (the target task, often with limited data). In the AD context, a deep learning model (e.g., a CNN) might be pre-trained on a massive dataset of general brain images (e.g., UK Biobank) or a large clinical database. The learned features and representations from this pre-training phase, which capture fundamental aspects of brain structure or clinical data patterns, can then be transferred to the specific AD prediction task. Typically, the initial layers of the pre-trained network are frozen (their weights are not updated), and only the final layers are fine-tuned using the smaller, task-specific AD dataset. This approach significantly reduces the amount of labeled data needed for the target task, accelerates the training process, and often leads to improved model performance and generalization by initializing the model with relevant, pre-learned feature extractors.

By strategically employing these diverse AI-driven techniques—ranging from generative modeling for data augmentation and balancing to advanced imputation and knowledge transfer—researchers can substantially alleviate the constraints imposed by common data challenges. Effectively addressing issues of dimensionality, sparsity, missingness, imbalance, and limited sample size is paramount for building the robust, accurate, and equitable personalized AD progression models required to advance clinical research and ultimately improve patient care.

\section{Discussion, Open Challenges, and Future Directions}
\label{sec:discussion}

The preceding sections have surveyed the rapidly evolving landscape of AI methodologies applied to the challenging task of personalized AD progression prediction. This review highlights a clear and accelerating trend away from traditional statistical methods and towards sophisticated computational approaches capable of harnessing the rich information contained within large-scale, multimodal, and longitudinal patient datasets. The overarching goal is to move beyond population-level staging or group-based prognostication towards generating accurate, individualized forecasts of cognitive and pathological trajectories, ultimately enabling more precise clinical decision-making, tailored interventions, and optimized clinical trial designs. This concluding section synthesizes the key findings, identifies persistent hurdles, and outlines promising avenues for future research in this critical domain.

\subsection{Discussion}
\label{subsec:discussion_synthesis}
The AI methodologies represent a diverse toolkit for modeling AD progression. SSMs offer a statistically rigorous framework for analyzing longitudinal data, adeptly handling measurement noise and within-subject correlations, making them particularly suitable for modeling trajectories based on relatively structured, lower-dimensional data like sequential cognitive scores or specific biomarker levels. However, their capacity to capture highly complex, non-linear dynamics inherent in high-dimensional data (e.g., raw neuroimaging) might be limited compared to deep learning counterparts. Deep learning temporal models, particularly RNNs and their variants like LSTM networks, excel at learning intricate temporal dependencies directly from sequential data, proving effective for forecasting based on multivariate time series integrating clinical, cognitive, and imaging information over time. Their strengths lie in automatic feature learning from complex sequences, but they often require substantial data, can be challenging to interpret (the "black box" problem), and may require careful handling of irregular sampling intervals common in clinical data. GNNs provide a unique advantage by explicitly modeling relational structures, whether representing functional or structural brain connectivity networks derived from neuroimaging or patient similarity graphs based on multimodal features. This allows GNNs to capture topological information potentially missed by other methods, although constructing meaningful graph representations and integrating temporal dynamics remain key considerations. Finally, the emerging paradigms of AI-driven digital twins and related deep generative models (like CRBMs) represent a significant conceptual leap, moving from prediction to simulation. These approaches hold immense promise for generating highly personalized forecasts under various hypothetical scenarios (e.g., effects of interventions), enabling in silico experimentation. However, they are computationally intensive, highly data-dependent, and their validation poses significant challenges.

The specific suitability of each method often depends on the nature of the available data and the prediction task. SSMs might be preferred for clearer interpretability with structured longitudinal data, while RNNs/LSTMs are powerful for complex sequences if sufficient data is available. GNNs are indispensable when network structure is paramount, and DTs offer the potential for the deepest level of personalization and simulation, albeit with higher complexity. A crucial implication evident throughout the reviewed literature is the undeniable shift towards leveraging both \textbf{multimodal data integration} and \textbf{longitudinal modeling}. Recognizing that AD is a complex systemic disease manifesting across multiple biological scales and evolving over time, models that synergistically combine diverse data sources (imaging, fluid biomarkers, genetics, cognition, demographics) and explicitly capture temporal dynamics consistently demonstrate superior performance in predicting individual trajectories compared to static, unimodal approaches.

Despite the impressive predictive capabilities of many advanced AI models, particularly deep learning architectures, their clinical translation is often hampered by concerns regarding \textbf{interpretability and generalizability}. Clinicians and regulatory bodies require not just accurate predictions but also an understanding of \textit{why} a model arrives at a specific forecast for a patient. While techniques like SHAP values, saliency maps, and attention mechanisms are increasingly employed to shed light on model decisions, achieving true clinical interpretability remains an active area of research. Equally critical is generalizability; models must perform reliably not just on the dataset they were trained on (e.g., ADNI) but also on unseen data from different patient populations, clinical settings, and acquisition protocols. Studies emphasizing external validation, often using diverse datasets like NACC, are essential for building trust and ensuring real-world applicability. The potential for models trained on relatively homogenous research cohorts to underperform or exhibit bias when applied to more diverse clinical populations is a significant concern that necessitates rigorous validation across varied demographic and clinical contexts.

\subsection{Open Challenges}
\label{subsec:challenges}
While AI has propelled personalized AD prediction forward, several significant challenges must be addressed to realize its full clinical potential:

\begin{itemize}
    \item \textbf{Robust External Validation and Benchmarking:} Many promising models are primarily validated on subsets of ADNI or similar research cohorts. There is a critical need for rigorous validation on large, diverse, independent, and prospectively collected clinical datasets reflecting real-world patient heterogeneity. Furthermore, the lack of standardized evaluation metrics and universally accepted benchmark datasets specifically designed for personalized \textit{longitudinal} prediction makes direct comparison between different modeling approaches difficult.
    \item \textbf{Clinical Workflow Integration:} Translating sophisticated AI models from research environments into practical tools usable by clinicians at the point of care remains a major hurdle. This involves addressing issues of usability, computational requirements, integration with existing electronic health record (EHR) systems, regulatory approval processes (e.g., FDA, EMA), and fostering clinician trust and acceptance.
    \item \textbf{Ethical Considerations:} The use of sensitive patient data for developing AI models raises significant ethical concerns. These include ensuring patient privacy and data security, mitigating algorithmic bias (ensuring models perform fairly across different demographic groups like race, sex, and socioeconomic status), establishing accountability for model predictions, and ensuring transparency in model development and deployment.
    \item \textbf{Modeling Therapeutic Interventions:} Most current models focus on predicting the natural history of AD progression. A major unmet need is the development of dynamic models capable of incorporating the effects of pharmacological treatments, lifestyle interventions (e.g., diet, exercise), or other clinical management strategies to predict individual responses and optimize personalized treatment plans.
    \item \textbf{Handling Comorbidities:} AD rarely occurs in isolation, especially in older adults. Patients often present with multiple comorbidities (e.g., vascular disease, diabetes, depression) that can significantly influence AD progression and presentation. Developing models that can effectively account for these complex interactions is crucial for accurate personalized prediction in real-world clinical scenarios.
    \item \textbf{Data Scarcity for Subtypes and Underrepresented Groups:} Despite large overall datasets, obtaining sufficient data for specific AD subtypes (e.g., early-onset AD, rare genetic forms) or demographic groups historically underrepresented in research remains a challenge, limiting the development of truly personalized models for these populations. Synthetic data generation (Section~\ref{sec:data_challenges}) offers partial solutions, but cannot fully replace real-world data.
\end{itemize}

\subsection{Future Directions}
\label{subsec:future_directions}
Addressing the open challenges requires concerted research efforts across multiple fronts. Promising future directions include:

\begin{itemize}
    \item \textbf{Development of Hybrid Models:} Combining the strengths of different AI paradigms could lead to more robust and interpretable models. Examples include integrating mechanistic or systems biology models with data-driven machine learning, using deep learning for feature extraction from high-dimensional data fed into statistically rigorous SSMs for trajectory modeling, or combining GNNs (for network structure) with RNNs (for temporal dynamics).
    \item \textbf{Causal Inference Methods:} Moving beyond predictive correlation towards understanding the causal drivers of heterogeneous AD progression is essential for developing effective interventions. Applying emerging AI techniques for causal discovery and inference to longitudinal AD data could yield crucial insights into patient-specific mechanisms of decline.
    \item \textbf{Federated Learning and Privacy-Preserving AI:} To overcome data access limitations and privacy concerns while enabling model training on larger, more diverse datasets, federated learning approaches, where models are trained locally at different institutions without sharing raw patient data, hold significant promise. Other privacy-enhancing technologies (e.g., differential privacy, homomorphic encryption) will also be crucial.
    \item \textbf{More Sophisticated Digital Twin Frameworks:} Future digital twins could become more dynamic, multi-scale, and mechanistic, integrating data from the molecular (omics) and cellular levels up to brain networks and clinical presentation. Incorporating agent-based modeling or more detailed physiological simulations could enhance their predictive power and utility for testing interventions in silico \cite{soykanWintersim2024}.
    \item \textbf{Real-Time Prediction Updates and Online Learning:} Developing models capable of continuously updating individual patient prognoses as new data (e.g., from wearable sensors, regular clinical visits) becomes available would significantly enhance their clinical utility for ongoing patient management and adaptive trial designs.
    \item \textbf{Integration of Novel and Digital Biomarkers:} Incorporating newly validated biomarkers (e.g., blood-based markers like p-tau217) and passively collected digital biomarkers (e.g., from smartphone usage patterns, speech analysis, wearable sensor data) into predictive models could provide richer, more continuous, and less invasive data streams for tracking progression.
    \item \textbf{Explainable AI (XAI) Tailored for Clinical Utility:} Advancing XAI techniques beyond generic feature importance measures towards generating explanations that are truly meaningful, actionable, and trustworthy for clinicians making decisions about individual patients is paramount for bridging the gap to clinical practice.
    \item \textbf{Community-Driven Benchmarking Efforts:} Establishing collaborative platforms and initiatives to create standardized, openly available benchmark datasets, clearly defined prediction tasks (e.g., predicting cognitive decline over specific horizons, time-to-event prediction), and transparent evaluation protocols would greatly facilitate progress and reproducibility in the field.
\end{itemize}

In conclusion, while significant challenges remain, the application of AI to personalized AD progression prediction holds immense potential. By continuing to develop innovative methodologies, addressing data limitations, prioritizing interpretability and validation, and fostering interdisciplinary collaboration, the field is poised to deliver tools that can transform AD research and care, leading towards a future of more precise, proactive, and personalized management of this devastating disease.

\section{Conclusion}
\label{sec:conclusion}

This paper has provided a comprehensive survey of the application of AI methodologies to the critical challenge of personalized AD progression prediction. The inherent heterogeneity in the clinical manifestation and temporal evolution of AD renders traditional, population-based prognostic approaches inadequate for individual patient management, highlighting an urgent need for tools capable of forecasting disease trajectories on a person-specific basis. This review charts the significant progress made by leveraging AI to address this need, moving towards more precise and individualized neurological care.

We have explored key AI techniques, ranging from established statistical frameworks like SSMs adapted for longitudinal analysis, to cutting-edge deep learning approaches including RNNs for capturing temporal dynamics and GNNs for modeling complex relationships within brain networks or patient cohorts. Also, we discussed the burgeoning field of AI-driven digital twins and generative models, which offer the potential not just to predict, but to simulate individual disease pathways under varying conditions. A common thread across these advanced methods is their ability to integrate complex, multimodal data sources—spanning neuroimaging, genetics, clinical assessments, and biomarkers—over time, thereby capturing a richer, more dynamic picture of individual disease processes than previously possible.

Despite the power of these AI tools, their development and application are often constrained by significant data challenges, including high dimensionality, sparsity and irregularity in longitudinal measurements, missing data points, and dataset imbalances reflecting varied progression rates or underrepresented subgroups. We reviewed how AI itself offers promising mitigation strategies, particularly through sophisticated imputation techniques and the generation of realistic synthetic data using VAEs and GANs, which can help augment datasets and address class imbalances.

Nevertheless, substantial hurdles remain before these AI-driven personalized prediction models can be fully integrated into routine clinical practice. Key open challenges include the need for rigorous external validation across diverse populations to ensure generalizability and fairness, enhancing model interpretability to foster clinical trust (Explainable AI), addressing profound ethical considerations regarding data privacy and algorithmic bias, integrating the effects of treatments into predictive models, and establishing standardized benchmarking protocols. Future research directions are focused on overcoming these obstacles through the development of hybrid models combining mechanistic insights with data-driven approaches, leveraging causal inference methods to understand progression drivers, employing federated learning for privacy-preserving multi-site collaboration, creating more sophisticated and dynamic digital twin frameworks, integrating novel biomarker data streams, and ensuring that AI tools are designed for seamless clinical workflow integration.

AI stands poised to fundamentally transform our ability to understand, predict, and manage AD at the individual level. By enabling accurate, personalized prognostication, AI holds the potential to facilitate earlier interventions, optimize treatment strategies, improve the design and efficiency of clinical trials, and ultimately alleviate the immense burden of AD on patients, families, and society. Continued innovation and diligent validation in this field promise a future where precision medicine becomes a reality for individuals affected by AD. 

\section*{Acknowledgment}

The bibliometric visualizations in this study were generated using VOSviewer\cite{van2010software}, an open-source software tool developed by the Centre for Science and Technology Studies at Leiden University. The use of VOSviewer is permitted under the Creative Commons Attribution 4.0 International License, and appropriate attribution has been provided accordingly.

\bibliographystyle{unsrt}  
\bibliography{references}

\end{document}